\documentclass[dvipsnames,format=sigconf,anonymous=false,review=false]{acmart}
\usepackage[algo2e,ruled,vlined,linesnumbered]{algorithm2e}
\usepackage[utf8]{inputenc}
\usepackage{multirow}
\usepackage{hyperref}
\usepackage{caption}
\usepackage{subcaption}
\hypersetup{
    colorlinks=true,
    linkcolor=blue,
    filecolor=magenta,      
    urlcolor=cyan,
}

\usepackage{xspace}
\newcommand{\OM}{\textsc{OneMax}\xspace}
\newcommand{\LO}{\textsc{LeadingOnes}\xspace}
\newcommand{\OMM}{\textsc{OneMinMax}\xspace}
\newcommand{\LOTZ}{\textsc{LOTZ}\xspace}
\newcommand{\OJZJ}{\textsc{OneJumpZeroJump}\xspace}
\newcommand{\COCZ}{\textsc{COCZ}\xspace}

\title{
Towards Self-adaptive Mutation in Evolutionary Multi-Objective Algorithms
}

\author{Furong Ye}
\email{f.ye@liacs.leidenuniv.nl}
\orcid{0000-0002-8707-4189}
\affiliation{%
  \institution{LIACS, Leiden University}
  \streetaddress{Snellius Building, Niels Bohrweg 1}
  \city{Leiden}
  \country{Netherlands}
  \postcode{2333 CA}
}
\author{Frank Neumann}
\email{frank.neumann@adelaide.edu.au}
\orcid{0000-0002-2721-3618}
\affiliation{%
  \institution{The University of Adelaide}
  \city{Adelaide}
  \country{Australia}
}
\author{Jacob de Nobel}
\email{j.p.de.nobel@liacs.leidenuniv.nl}
\orcid{0000-0003-1169-1962}
\affiliation{%
  \institution{LIACS, Leiden University}
  \streetaddress{Snellius Building, Niels Bohrweg 1}
  \city{Leiden}
  \country{Netherlands}
  \postcode{2333 CA}
}
\author{Aneta Neumann}
\email{aneta.neumann@adelaide.edu.au}
\orcid{0000-0002-0036-4782}
\affiliation{%
  \institution{The University of Adelaide}
  \city{Adelaide}
  \country{Australia}
}

\author{Thomas B\"ack}
\email{T.H.W.Baeck@liacs.leidenuniv.nl}
\orcid{0000-0001-6768-1478}
\affiliation{%
  \institution{LIACS, Leiden University}
  \streetaddress{Snellius Building, Niels Bohrweg 1}
  \city{Leiden}
  \country{Netherlands}
  \postcode{2333 CA}
}

\begin{document}

\copyrightyear{2023}
\acmYear{2023}
\setcopyright{acmlicensed}
\acmConference[FOGA '23]{Foundations of Genetic Algorithms}{August 30--September 1, 2023}{Potsdam, Germany}
\acmBooktitle{Foundations of Genetic Algorithms (FOGA '23), August 30--September 1, 2023, Potsdam, Germany}
\acmPrice{15.00}
\acmDOI{xxx}
\acmISBN{xxx}

\begin{abstract}
Parameter control has succeeded in accelerating the convergence process of evolutionary algorithms.
While empirical and theoretical studies have shed light on the behavior of algorithms for single-objective optimization, little is known about how self-adaptation influences multi-objective evolutionary algorithms. In this work, we contribute (1) extensive experimental analysis of the Global Simple Evolutionary Multi-objective Algorithm (GSEMO) variants on classic problems, such as \OMM, \LOTZ, \COCZ, and (2) a novel version of GSEMO with self-adaptive mutation. 

To enable self-adaptation in GSEMO, we explore three self-adaptive mutation techniques from single-objective optimization and use various performance metrics, such as hypervolume and inverted generational distance, to guide the adaptation. Our experiments show that adapting the mutation rate based on single-objective optimization and hypervolume can speed up the convergence of GSEMO. Moreover, we propose a GSEMO with self-adaptive mutation, which considers optimizing for single objectives and adjusts the mutation rate for each solution individually. Our results demonstrate that the proposed method outperforms the GSEMO with static mutation rates across all the tested problems.

This work provides a comprehensive benchmarking study for MOEAs and complements existing theoretical runtime analysis. Our proposed algorithm addresses interesting issues for designing MOEAs for future practical applications. 

\end{abstract}
\keywords{Multi-objective evolutionary algorithm, self-adaptation, mutation, benchmarking}

\maketitle

\section{Introduction}
Evolutionary algorithms (EAs) can find global optima by creating offspring solutions through global search variators. However, in practical applications, we can not afford infinite running time in real-world applications, and the convergence rate of EAs is a major concern. To address this issue, parameter control techniques have been developed to accelerate the convergence speed of EAs. Understanding the relationship between variator parameters and the dynamic behavior of algorithms is essential to improve their performances.  
For example, in the context of single-objective pseudo-boolean optimization $f: \{0,1\}^n \rightarrow \mathbb{R}$, EAs create offspring $y$ by flipping $0 < \ell \leq n$ bits of the parent solution $x$, where $\ell$ can be sampled from different distributions depending on the design of mutation operators. Previous studies~\cite{DBLP:journals/tec/EibenHM99,DBLP:journals/algorithmica/DoerrD18,DBLP:conf/gecco/DangD19,DBLP:journals/algorithmica/DoerrGWY19} have shown the choice of $\ell$ has a significant impact on the performance of EAs for the classic problems such as \OM and \LO, and self-adaptive methods have demonstrated higher convergence rates than the methods that use static settings. 

While there have been detailed studies on parameter control in the single-objective optimization domain, corresponding investigations for multi-objective benchmark problems are lacking in the literature. Regarding runtime analysis, only a few classic multi-objective benchmark problems such as \OMM, \LOTZ, and \COCZ have been rigorously studied. Recently, a multi-modal multi-objective benchmark problem \OJZJ has been introduced. Investigations in the area of runtime analysis have been started by \cite{DBLP:journals/tec/LaumannsTZ04}, where the authors studied a variant of the simple evolutionary multi-objective optimizer (SEMO) which produces an offspring by flipping a single bit and always maintaining a set of non-dominated solutions according to the given objective functions. Runtime bounds of $\Theta(n^3)$ for LOTZ and $O(n^2 \log n)$ for COCZ have been shown in \cite{DBLP:journals/tec/LaumannsTZ04}. \OMM has been investigated in \cite{DBLP:journals/ec/GielL10,DBLP:journals/tcs/NguyenSN15,DBLP:conf/gecco/DoerrGN16} and it has been shown that the global simple evolutionary multi-objective optimizer (GSEMO), which differs from SEMO by applying standard bit mutations instead of single bit flips, computes the whole Pareto front for \OMM in expected time $O(n^2 \log n)$. Furthermore, hypervolume-based evolutionary algorithms have been studied for \OMM in \cite{DBLP:journals/tcs/NguyenSN15,DBLP:conf/gecco/DoerrGN16} and the computation of structural diverse populations has been investigated in \cite{DBLP:conf/gecco/DoerrGN16}. In~\cite{DBLP:journals/tcs/OsunaGNS20}, different parent selection methods have been analyzed for \OMM and \LOTZ and their benefit has been shown when incorporated into GSEMO. Recently, \OMM has also been used to study the runtime behavior of the NSGA-II algorithm~\cite{DBLP:conf/aaai/0001LD22}. All the bounds obtained are asymptotic, i.e. they are missing the leading constants. Therefore, it is interesting to carry out more detailed experimental investigations of simple evolutionary multi-objective algorithms from the area of runtime analysis alongside some variations, such as different mutation operators and the use of larger offspring population sizes.

In this work, we focus on the mutation operators and explore self-adaptation to accelerate the convergence speed of multi-objective evolutionary algorithms (MOEAs). Our detailed experimental investigation provides complementary insights that facilitate progress in the theoretical understanding of evolutionary multi-objective optimization. Furthermore, the proposed MOEA using self-adaptive mutation exhibits competitive results, which may inspire the design of MOEAs for practical applications. 

Specifically, we equip GSEMO with self-adaptive mutation operators designed for single-objective optimization, and present results on \OMM, \LOTZ, and \COCZ. 
As each solution in multi-objective optimization $f: \mathbb{R}^n \rightarrow \mathbb{R}^m$, where $m$ is the number of the objectives (note that we consider only the bi-objective case, i.e., $m=2$, in this work), maps to a set of objective function values, we use performance metrics such as hypervolume, inverted generational distance, and a single-objective oriented metric to guide adaptation techniques for GSEMO. The results show that the self-adaptive techniques can improve the convergence speed of GSEMO. Following our analysis, we propose a GSEMO with a self-adaptive mutation rate that outperforms GSEMO using a static mutation rate across all tested problems.

The study not only provides guidelines for designing self-adaptive operators for MOEAs but also complements the existing theoretical analysis of multi-objective optimization by demonstrating the convergence process of GSEMO. As such, we hope this detailed experimental investigation inspires further progress in the theoretical understanding of evolutionary multi-objective optimization. 

\section{Preliminaries}
\subsection{Benchmark Problems}
\label{sec:bench}
In this paper, we focus on three classic multi-objective optimization problems that have been extensively studied in the field of runtime analysis. Below are the problem definitions for these benchmarks.
\subsubsection{\OMM}
\OMM introduced in \cite{DBLP:journals/ec/GielL10} is a bi-objective pseudo-Boolean optimization problem that generalizes the classical single-objective OneMax problem to the bi-objective case:
\begin{equation}
    \OMM :  \{0,1\}^n \rightarrow \mathbb{N}^2, x \mapsto \left (\sum_{i=1}^n x_i, n - \sum_{i=1}^n x_i \right )
\end{equation}
The problem is maximizing the numbers of both \emph{one} and \emph{zero} bits, and the objective of maximizing the number of \emph{one} bits is identical to the classic pseudo-Boolean optimization problem $\textsc{OneMax}: \{0,1\}^n \rightarrow \mathbb{N},  x \mapsto \sum_{i=1}^n x_i$. For the \OMM problem, all solutions locate at the optimal Parent front, and the goal is to obtain the entire set of Pareto front $\{(i,n-1)\mid i \in \{0,\ldots,n\}\}$.

\subsubsection{\LOTZ}
The \emph{Leading Ones, Trailing Zeroes} (\LOTZ) introduced in \cite{DBLP:journals/tec/LaumannsTZ04} maximizes the numbers of leading \emph{one} bits and trailing \emph{zero} bits, simultaneously. The problem can be defined as
\begin{equation}
    \LOTZ: \{0,1\}^n \rightarrow \mathbb{N}^2, x \mapsto \left (\sum_{i=1}^n \prod_{j=1}^i x_j, \sum_{i=1}^n \prod_{j=i}^n (1-x_j) \right )
\end{equation}
The Pareto front of \LOTZ~is $ \{(i,n-i) \mid i \in \{0,\ldots,n\}\}$, given by the set of $n+1$ solutions $x = \{1^i0^{(n-i)}\mid i \in \{0,\ldots,n\}\}$. One property of \LOTZ is that for all non-dominated solutions, the neighbors that are with $\textbf{1}$ Hamming distance are either better or worse but incomparable.

\subsubsection{\COCZ} \emph{Count Ones Count Zeroes}~\cite{DBLP:journals/tec/LaumannsTZ04} is another extension of the \OM problem and defined by:
\begin{equation}
     \COCZ :  \{0,1\}^n \rightarrow \mathbb{N}^2, x \mapsto \left (\sum_{i=1}^n x_i, \sum_{i=1}^{n/2} x_i + \sum_{i=n/2}^{n} (1-x_i) \right )
\end{equation}
where $n = 2k, k \in \mathbb{N}$. The Pareto front of \COCZ is a set $\mathcal{P}$ consisting of the solutions with $n/2$ ones in the first half of the bit string. The size of $\mathcal{P}$ is $n/2$.
Differently from \OMM, of which all possible solutions locate at the Pareto front, many solutions that are strictly dominated by others exist in the search space of \COCZ.

\subsection{The GSEMO Algorithm}
The simple evolutionary multi-objective optimizer (SEMO) \cite{DBLP:journals/tec/LaumannsTZ04} uses a population $P$ of non-dominated solutions. The population is initialized with a random solution. Then, the algorithm creates offspring by selecting a solution $x$ from $P$ uniformly at random (u.a.r.) and flipping one bit of $x$. If any solutions of $P$ are dominated by $x$, those solutions will be removed, and $x$ will be added to $P$. The algorithm runs until reaching a termination condition, e.g., the budget is used out, or the entire Parent front is found. Global SEMO (GSEMO)~\cite{1299908} differs from SEMO by applying the \emph{standard bit mutation} instead of flipping exactly \emph{one-bit} when creating offspring. As shown in Algorithm~\ref{alg:GSEMO}, GSEMO applies standard bit mutation which flips \emph{at least} one bit. More precisely, $\ell$, the number of flipped bits, is sampled from a conditional binomial distribution $\text{Bin}_{>0}(n,p)$ \cite{DBLP:journals/corr/abs-1812-00493}, where $p$ is the mutation rate and $n$ the dimensionality. Offspring is then created by $\text{flip}_\ell (x)$ which flips $\ell$ bits of $x$ chosen u.a.r. This version of GSEMO, which uses a fixed mutation rate of $p=1/n$, we name static GSEMO throughout this paper. 

\begin{algorithm2e}
\caption{Global SEMO}
\label{alg:GSEMO}
\textbf{Input:} mutation rate $p = 1/n$\;
\textbf{Initialization:} Sample $x \in \{0,1\}^n$ u.a.r., and evaluate $f(x)$\;
$P \leftarrow \{x\}$\;
\textbf{Optimization:} \While{not stop condition}{
Select $x \in P$ u.a.r.\;
Sample $\ell \sim \text{Bin}_{>0}(n,p)$, create $y \leftarrow \text{flip}_{\ell}(x)$, and evaluate $f(y)$\;
\If{there is no $z\in P$ such that $y \preceq z$}{$P = \{z \in P \mid z \not\preceq y\} \cup \{y\}$}
}
\end{algorithm2e}

\section{Translating Single-Objective Techniques}  
\label{sec:multi-adap}
\subsection{The Self-Adaptive GSEMO Variants}
As demonstrated in previous studies \cite{DBLP:journals/tec/EibenHM99,DBLP:journals/algorithmica/DoerrD18,DBLP:conf/gecco/DangD19,DBLP:journals/algorithmica/DoerrGWY19},
the optimal parameter settings of EAs can change during the optimization process. Empirical and theoretical results~\cite{DBLP:journals/algorithmica/DoerrGWY19,DBLP:journals/algorithmica/DoerrD18,DBLP:conf/gecco/KruisselbrinkLREB11,DBLP:conf/cec/YeDB19} have shown that EAs with self-adaptive mutations can outperform those with standard bit mutation for classical single-objective problems such as \OM, \LO, and \textsc{Jump}. 

Self-adaptation techniques for MOEAs have been studied to a lesser degree. 
An algorithm called MO $(\mu+(\lambda, \lambda))$~GA, which translates the $(1+(\lambda, \lambda))$~GA from \cite{DBLP:journals/tcs/DoerrDE15} to the multi-objective setting, has been introduced in \cite{DBLP:journals/algorithmica/ShiSFKN19} for solving \OM under dynamic uniform constraints.
More recently, a algorithm called the $(1+(\lambda,\lambda))$~GSEMO was proposed in~\cite{DBLP:conf/gecco/DoerrHP22}, which showed that adapting the population size for $(1+(\lambda,\lambda))$~GSEMO reduces the expected optimization time for OneMinMax to $O(n^2)$.
Furthermore, a self-adjusting GSEMO~\cite{DBLP:conf/aaai/DoerrZ21}, which increases the mutation rate after $T$ iterations of not obtaining new non-dominated solutions, was tested for \OJZJ. 
In this section, we test three GSEMO variants with self-adaptive mutation operators that have been extensively studied in single-objective benchmarking~\cite{DBLP:journals/asc/DoerrYHWSB20}. We focus on the self-adaptive mutation operators and investigate their effectiveness in the context of multi-objective optimization. In our implementation, all GSEMO variants sample an offspring population of size $\lambda$ in each generation, and the sampling distribution of $\ell$ is adjusted based on the value of $e(x)$ of the newly created solutions. The detailed procedures of the three GSEMO variants are introduced below, and the definitions of $e(x)$ will be provided later in this paper.

\subsubsection{The Two-rate GSEMO}
The two-rate EA with a self-adaptive mutation rate was proposed in~\cite{DBLP:journals/algorithmica/DoerrGWY19} and analyzed for \OM. The algorithm starts with an initial mutation rate of $r/n$, where $r$ is the expected mutation strength. In each generation, it samples 50\% of the offspring using a doubled mutation rate $(2r)/n$ and samples the other 50\% using a halved mutation rate $r/(2n)$. The mutation rate used to create the best offspring in this generation has a higher probability of $3/4$ being chosen for the next generation. The two-rate GSEMO (see Algorithm~\ref{alg:two-rate}) applies a $(1+\lambda)$~schema and samples offspring using the self-adaptive mutation scheme of the two-rate EA. To compare solutions' contributions $e(y)$ to the current non-dominated solution set $P$ (line 11), the algorithm uses the metrics that will be described in Section~\ref{sec:exp}. Note that we cap the mutation strength $r$ within $[1/2,n/4]$.

\begin{algorithm2e}
\caption{The two-rate GSEMO}
\textbf{Input:} Population size $\lambda$, $r^\text{init}$\; 
\textbf{Initialization:} Sample $x \in \{0,1\}^n$ u.a.r., and evaluate $f(x)$\;
$r \leftarrow r^\text{init}$ , $P \leftarrow \{x\}$\;
\textbf{Optimization:} \While{not stop condition}{
\For{i = 1,\ldots,$\lambda$}{
    Select $x \in P$ u.a.r.\;
    \eIf{$i < \lfloor \lambda/2 \rfloor$}{Sample $\ell^{(i)} \sim \text{Bin}_{>0}(n,r/(2n))$}{Sample $\ell^{(i)} \sim \text{Bin}_{>0}(n,(2r)/n)$,}
    Create $y^{(i)} \leftarrow \text{flip}_{\ell^{(i)}}(x)$, evaluate $f(y^{(i)})$,and assess $e(y^{(i)})$;
}
$y^{(i^*)} \leftarrow \arg\max\{e(y^{(1)}),\ldots,e(y^{(\lambda)})\}$\;
\lIf{$i^* < \lfloor \lambda/2 \rfloor$}{$s\leftarrow 3/4$ \textbf{ else } {$s \leftarrow 1/4$}}

Sample $q \in [0,1]$ u.a.r.\;
\lIf{$q \le s$}{$r \leftarrow \max\{r/2,1/2\}$ \textbf{else} $r \leftarrow \min\{2r, n/4\}$}
\For{$i = 1,\ldots,\lambda$}{
\If{there is no $z\in P$ such that $y^{(i)} \preceq z$}{$P = \{z \in P \mid z \not\preceq y^{(i)}\} \cup \{y^{(i)}\}$}
}}
\end{algorithm2e}

\subsubsection{The log-Normal GSEMO}
The log-normal GSEMO applies standard bit mutation and adjusts the mutation rate $p$ using a log-normal update rule~\cite{DBLP:conf/gecco/KruisselbrinkLREB11}. When creating a new offspring population, a mutation rate $p'$ is sampled for each new offspring as shown in line 7 of Algorithm~\ref{alg:lognormal}.
The log-normal strategy allows the mutation rate to increase and decrease with identical probabilities, and it chooses the $p'$ that is used to create the best solution as the $p$ for the next generation. We cap $p$ within $[1/4n,1/2]$.

\begin{algorithm2e}
\caption{The log-normal GSEMO}
\label{alg:lognormal}
\textbf{Input:} Population size $\lambda$, mutation rate $p$\; 
\textbf{Initialization:} Sample $x \in \{0,1\}^n$ u.a.r., and evaluate $f(x)$\;
$P \leftarrow \{x\}$\;
\textbf{Optimization:} \While{not stop condition}{
\For{$i = 1,\ldots,\lambda$}{
    Select $x \in P$ u.a.r.\;
    $p^{(i)} = \big(1+\frac{1-p}{p}\cdot \exp(0.22\cdot \mathcal{N}(0,1))\big)^{-1}$ \;
	Sample $\ell^{(i)} \sim \text{Bin}_{>0}(n,p^{(i)})$\;
	create $y^{(i)} \leftarrow \text{flip}_{\ell^{(i)}}(x)$, evaluate $f(y^{(i)})$,and assess $e(y^{(i)})$;
}
$y^{(i^*)} \leftarrow \arg\max\{e(y^{(1)}),\ldots,e(y^{(\lambda)})\}$\;
$p \leftarrow p^{(i^*)}$\;
\For{$i = 1,\ldots,\lambda$}{
\If{there is no $z\in P$ such that $y^{(i)} \preceq z$}{$P = \{z \in P \mid z \not\preceq y^{(i)}\} \cup \{y^{(i)}\}$}
}
}
\end{algorithm2e}

\subsubsection{The Variance Controlled GSEMO}
The variance-controlled GSEMO (var-ctrl) applies normalized bit mutation~\cite{DBLP:conf/cec/YeDB19} which samples $\ell$ from a normal distribution $N(r,\sigma^2)$ (line 6 in Algorithm~\ref{alg:normea}). Similar to the log-normal GSEMO, it also uses a greedy strategy and adjusts $r$ by the value of $\ell$ that is used to create the best solution. An advantage of using normal distributions is that we can control the mean and the variance of sampling $\ell$. In this work, we follow the setting in~\cite{DBLP:conf/cec/YeDB19} reducing the variance by a factor $F=0.98$ if $r$ remains the same in subsequent generations.

\begin{algorithm2e}[t]
\caption{The var-ctrl GSEMO}
\label{alg:normea}
\label{alg:two-rate}
\textbf{Input:} Population size $\lambda$, $r^\text{init}$\, and a factor $F$; 
\textbf{Initialization:} Sample $x \in \{0,1\}^n$ u.a.r., and evaluate $f(x)$\;
$r \leftarrow r^\text{init}$; $c \leftarrow 0$; $P \leftarrow \{x\}$\;
\textbf{Optimization:} \While{not stop condition}{
\For{$i = 1,\ldots,\lambda$}{
    Select $x \in P$ u.a.r.\;
    Sample $\ell^{(i)} \sim \min\{N_{>0}(r,F^cr(1-r/n)),n\}$\;
    create $y^{(i)} \leftarrow \text{flip}_{\ell^{(i)}}(x)$, evaluate $f(y^{(i)})$,and assess $e(y^{(i)})$;
}
$y^{(i^*)} \leftarrow \arg\max\{e(y^{(1)}),\ldots,e(y^{(\lambda)})\}$\;
\lIf{$r=\ell^{(i^*)}$}{$c\leftarrow c+1$ \textbf{else} $c\leftarrow 0$}
$r \leftarrow \ell^{(i^*)}$\;
\For{$i = 1,\ldots,\lambda$}{
\If{there is no $z\in P$ such that $y^{(i)} \preceq z$}{$P = \{z \in P \mid z \not\preceq y^{(i)}\} \cup \{y^{(i)}\}$}
}
}
\end{algorithm2e}

\subsection{Experiments}
\label{sec:exp}
\subsubsection{Performance Measures/Indicators}
Recall that the GSEMO variants rely on the value of $e(x)$ for self-adaption. For the single objective case, we could set $e(x)$ to the objective value $f(x)$, but for the multi-objective case, we need an alternative method for measuring the progress of the newly created solutions.
For multi-objective problems, in practice, given a population $P$ of the current non-dominated solutions, we measure the performance of a new solution $x$ based on the set $P \cup \{x\}$. Here, we apply two commonly used multi-objective performance measures and a single-objective oriented metric:

\begin{itemize}
    \item Hypervolume (HV) \cite{DBLP:journals/tec/ZitzlerT99}: The hypervolume indicator $I_H(P)$ is the volume of the objective space
    that is dominated by the solution set $P$. Given a reference point $s \in \mathbb{R}^m$, $I_H(P) = \Lambda \left( \bigcup_{x \in P}[f_1(x), s_1] \times \ldots \times [f_m(x), s_m] \right)$, where $\Lambda (P)$ is the Lebesgue measure of $P$ and $[f_1(x), s_1] \times \ldots \times [f_m(x), s_m]$ is the orthotope with $f(x)$ and $r$ in opposite corners. This paper calculates HV regarding the reference point $(-1,-1)$.
    \item Inverted generational distance (IGD) \cite{zitzler2003performance}: The generational distance (GD) measures how far the obtained non-dominated set $P$ is from the true Pareto front. It is defined as $\text{GD}(P) = \sqrt{\sum{d_i^2}} / |P|$, where $d_i$ is the Euclidean distance between the $i$-th solution in $P$ and its nearest solution in $\mathcal{P}$. Following the suggestion in~\cite{zitzler2003performance}, we apply IGD using the obtained non-dominated set $P$ as the reference set and calculate the distances from each solution of the Pareto front $\mathcal{P}$ to its nearest neighbor in $P$. In this way, the dynamic size of $P$ has less influence on the value of IGD.
    \item Single-objective oriented metric (OneObj\%$T$): In the process of GSEMO, we define $e(x)$ by either $(y_1 - y_1^*)$ or $(y_2 - y_2^*)$, where $y_1$ and $y_2$ are the two objective values of $x$, and $y_1^*$ and $y_2^*$ are the best corresponding objective values of the current non-dominated solution set. We sample a value $rand \in [0,1]$ u.a.r. every $T$ generation. When $rand < 0.5$, $e(x) = (y_1 - y_1^*)$; otherwise, $e(x) = (y_2 - y_2^*)$. We test $T\in \{1,10,50\}$, namely OneObj, OneObj\%10, and OneObj\%50.
\end{itemize}

It is worth mentioning that IGD is an \emph{accuracy} metric referring to the convergence to the full Pareto front $\mathcal{P}$, and both HV and IGD are \emph{diversity} metrics referring to the range of search space that is covered by the obtained non-dominated set $P$. The value of HV is based on a given reference set instead of the Parent front.

\subsubsection{Experimental setup}
We test the algorithms for the $100$ dimensional \OMM, \COCZ, and \LOTZ. For fairness of comparison, the static GSEMO in Algorithm~\ref{alg:GSEMO} also applies the $(\mu+\lambda)$ strategy, where $\mu$ is determined by the current size of the non-dominated solution set. The presented results are from $100$ independent runs.
For reproducibility, we provide the code on GitHub~\cite{codes} and data on Zenodo~\cite{zenodo}.

\begin{table*}[t]
    \centering

\begin{tabular}{rrrrrrr}
\toprule
 &   &         HV &        IGD &  OneObj &  OneObj\%10 &  OneObj\%50 \\
\midrule
\small{\OMM}& two-rate   &  $\mathbf{61\,624}$ &  $62\,039$ & $102\,509$  &     $104\,716$ &      $97\,240$ \\
&  & \textbf{(3.271e+08)} & (3.427e+08) & (8.226e+08) &  (1.206e+09) &  (7.451e+08)\\
\emph{static} = $68\,375$  & log-normal & $134\,859$ & $143\,533$ &  $78\,679$  &      $80\,495$ &     $83\,078$ \\
(4.208e+08)& & (2.201e+09) & (3.310e+09) & (6.016e+08) &  (7.462e+08) &  (1.165e+09) \\
& var-ctrl   & $205\,175$ & $214\,801$ & $135\,643$ &     $136\,596$ &     $130\,143$ \\
& & (5.840e+09) & (5.697e+09) & (1.950e+09) &  (2.188e+09) &  (1.688e+09) \\
\midrule

\small{\LOTZ}& two-rate   & $370\,956$ & $352\,083$ &  $409\,658$ &     $405\,611$ &     $423\,125$ \\
& & (6.602e+09) & (6.209e+09) & (8.429e+09) &  (8.751e+09) &  (1.058e+10) \\
\emph{static} = $340\,072$&  log-normal & $374\,231$ &    $728\,462$ &  $318\,825$ &     $332\,063$ &     $341\,046$ \\
(5.946e+09)& & (9.614e+09) &       (3.040e+10)  & (5.537e+09) &  (5.896e+09) &  (6.458e+09) \\
& var-ctrl   &  $432\,273$ &    $944\,021$ &  $272\,628$ &     $284\,322$ &     $\mathbf{263\,883}$ \\
&  & (2.913e+10) &       (5.925e+10) & \textbf{(3.770e+09)} &  (4.695e+09) &  (4.036e+09) \\

\midrule

\small{\COCZ} & two-rate   & $\mathbf{26\,921}$ &  $27\,036$ &   $38\,327$ &      $41\,654$ &      $40\,129$ \\
&  & (1.131e+08) & \textbf{(6.336e+07)} & (1.659e+08) &  (1.467e+08) &  (2.018e+08) \\
\emph{static} = $30\,049$&  log-normal & $31\,043$ &  $66\,068$ &   $32\,029$ &      $31\,855$ &      $31\,248$ \\
(1.217e+08)& & (1.216e+08) & (7.427e+08) & (2.048e+08) &  (1.525e+08) &  (1.052e+08) \\
& var-ctrl   & $46\,091$ & $109\,648$ &   $48\,414$ &      $50\,671$ &      $49\,310$ \\
& & (2.251e+08) & (1.535e+09) & (3.518e+08) &  (3.223e+08) &  (4.444e+08) \\
\bottomrule
\end{tabular}

    \caption{The mean and variance (in brackets) of function evaluations that each algorithm (listed in the second column) using the corresponding metric (listed in the first row) uses to obtain the entire Parent front of the $100$-dimensional \OMM, \COCZ, and \LOTZ. The results are average of $100$ runs. $\lambda = 10$. The results of the static GSEMO with static $p=1/n$ are listed in the first column. Bold values indicate the best result for each problem.}
    \vspace{-0.3cm}
    \label{tab:FE}
\end{table*}

\begin{figure*}[t]
    \centering
    \includegraphics[width=\linewidth]{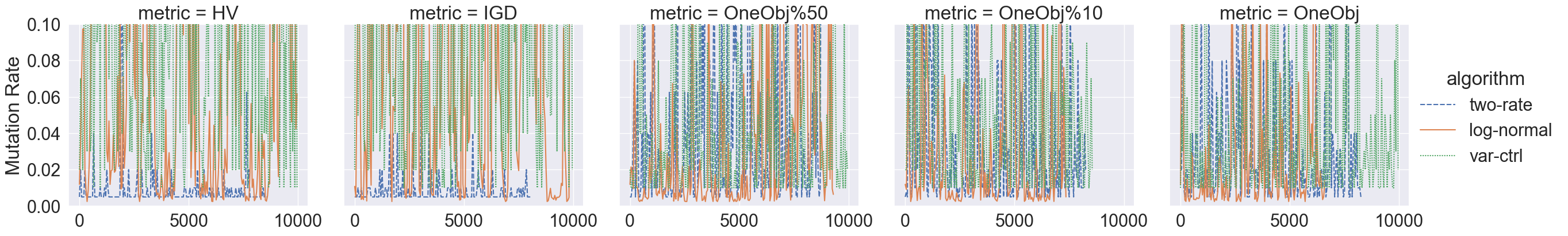}
    \includegraphics[width=\linewidth]{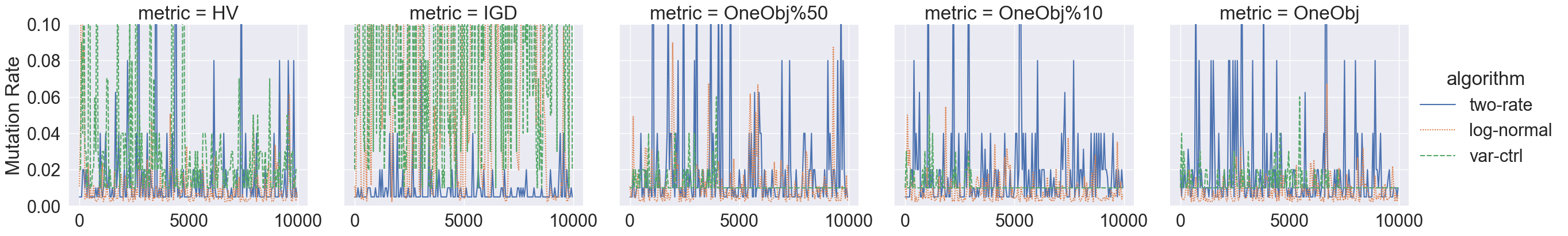}
    \includegraphics[width=\linewidth]{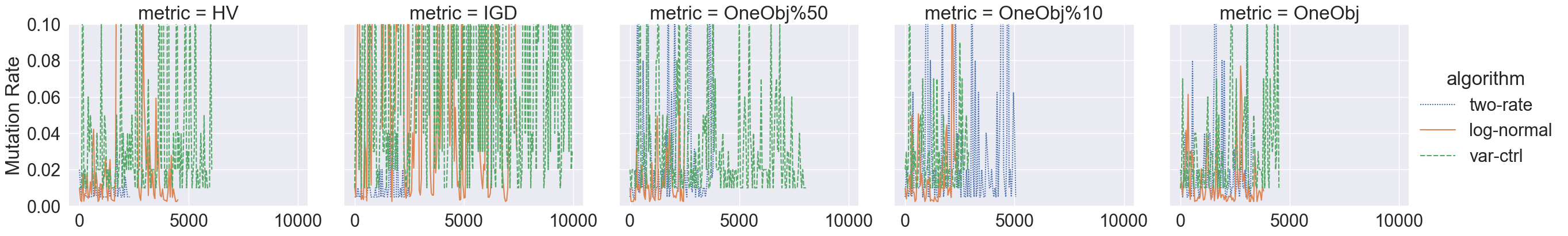}
    \caption{The self-adaptive mutation rates of the GSEMO variants during the optimization process. The results are from the GSEMO variants using five metrics for $100$-dimensional \OMM (\textbf{top}), \LOTZ (\textbf{mid}), and \COCZ (\textbf{bottom}). The plotted values are the exact values of one run for each algorithm. For ease of visualization, we plot the values at every $50$ generation and cap the maximal generation by $10\,000$ and mutation rates by $0.1$.}
    \vspace{1cm}
    \label{fig:MR}
\end{figure*}

\subsubsection{Running Time Results}
Table~\ref{tab:FE} presents the average function evaluations (FEs) for the various GSEMO algorithms to obtain the entire Pareto front. While self-adaptation can provide advantages over the static GSEMO, no algorithm outperforms the others across all the tested problems, and the choice of cost metric $e(x)$ clearly impacts the results. In practice, the two-rate GSEMO using HV performs the best for \OMM, while the var-ctrl GSEMOs using OneObj\%50 and using OneObj are the best ones for \LOTZ. The former achieves the best FEs, and the latter has a smaller variance. For \COCZ, the two-rate GSEMOs using HV and IGD show the lowest number of FEs and variance, respectively. Overall, this shows that \emph{self-adaptation can help obtain the entire Pareto front faster,} but proper design and cost metric selection for self-adaptive mutations are necessary.

To understand the (dis)advantages of the GSEMO variants' performance, we plot in Figure~\ref{fig:MR} adaptive mutation rates during the optimization process. For \OMM, mutation rates of the two-rate GSEMO are generally smaller than $0.05$ when using HV and IGD, while mutation rates of the other algorithms fluctuate and sometimes exceed $0.1$. Based on Table~\ref{tab:FE}, the two-rate GSEMO achieves better results by adjusting the mutation rate in the range of small values. The two-rate strategy adjusts the mutation rate following increasing (i.e., double) and decreasing (i.e., half) trends, but the log-normal and var-ctrl strategies adjust it by sampling values from a log-normal or normal distribution. Figure~\ref{fig:MR}
shows that the latter ones are inhibited by sampling and selecting large mutation rates. We infer that the algorithms can make progress using very large mutation rates in the early optimization process of \OMM, although optimal values are relatively small. For \COCZ, we observe similar results to \OMM. However, for \LOTZ, mutation rates of the log-normal and var-ctrl GSEMO fluctuate among smaller values and obtain promising results, particularly with objective-oriented metrics. It is important to note that while optimizing \OMM, significant progress can be made in the early stages by creating novel solutions using large mutation rates. In contrast, for \LOTZ, progress can be accomplished only when the number of leading ones or trailing zeros increases, and this requires relatively small mutation rates.

Additionally, we conduct a Mann-Whitney U rank test on the function evaluations from $100$ independent runs of the best-performing algorithms compared to the static GSEMO. Specifically, we test the two-rate GSEMO using HV for \OMM, the var-ctrl GSEMO using OneObj\%50 for \LOTZ, and the two-rate GSEMO using HV for \COCZ. The $p$-values are $0.013$, 7.904e-12, and $0.016$ resp.

\begin{figure}
    \centering
    \begin{subfigure}[b]{.44\textwidth}
    \includegraphics[width=\linewidth]{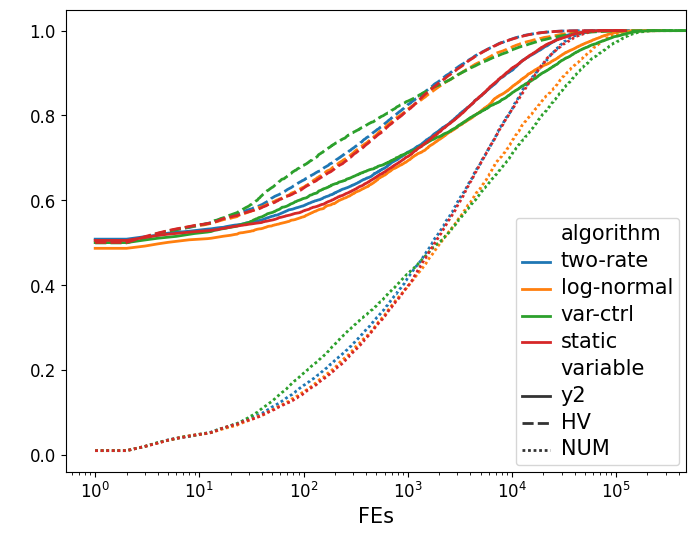}
    \vspace{-0.5cm}
    \caption{\OMM \& Using HV}

    \end{subfigure}
    \hfill
   \begin{subfigure}[b]{.44\textwidth}
    \includegraphics[width=\linewidth]{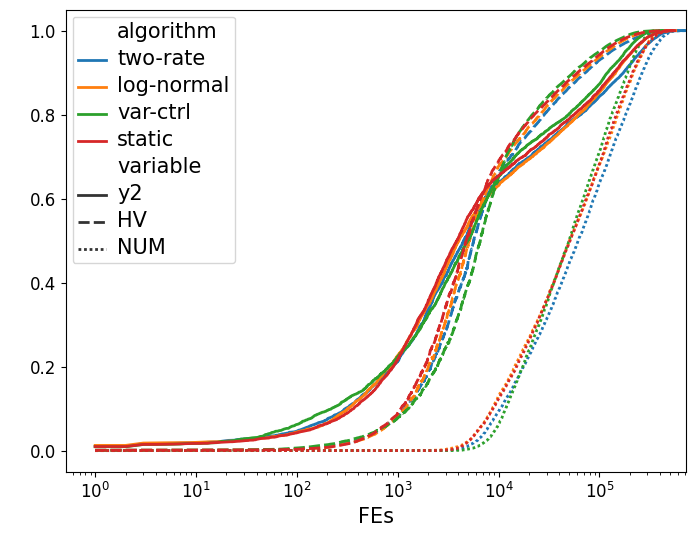}
        \vspace{-0.5cm}
     \caption{\LOTZ \& Using OneObj}
     \end{subfigure}
    \hfill
     \begin{subfigure}[b]{.44\textwidth}
    \includegraphics[width=\linewidth]{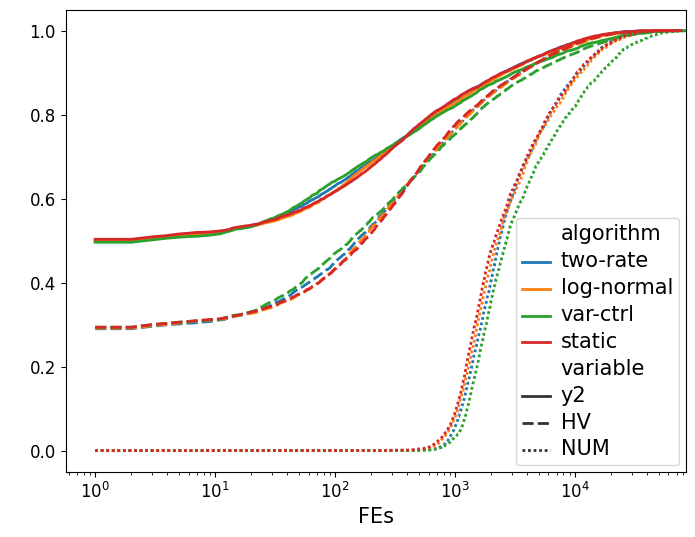}
    \vspace{-0.5cm}
     \caption{\COCZ \& Using HV}
     \vspace{-0.3cm}
     \end{subfigure}
    \caption{The convergence process of $y2$, Hypervolume (HV), and the number of obtained Pareto solutions (NUM) for $100$-dimensional \OMM, \LOTZ, and \COCZ.
   Plotted values are normalized by the corresponding maximum.
    Results are the average of $100$ runs.}
    \label{fig:variables}
\end{figure}

\begin{figure}
    \centering
    \includegraphics[width=.48\linewidth]{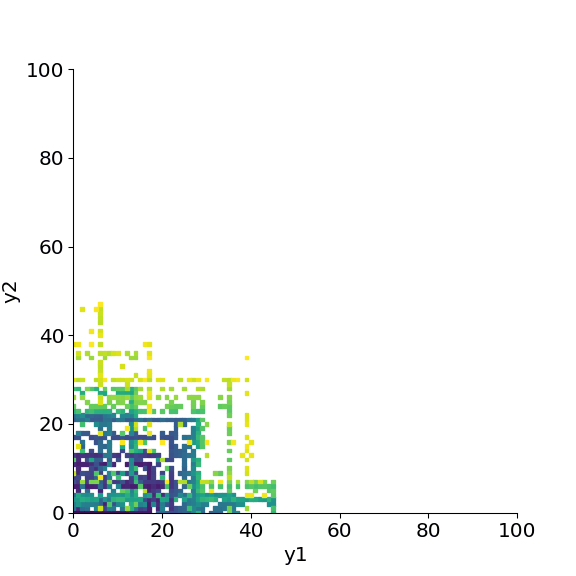}
    \includegraphics[width=.48\linewidth]{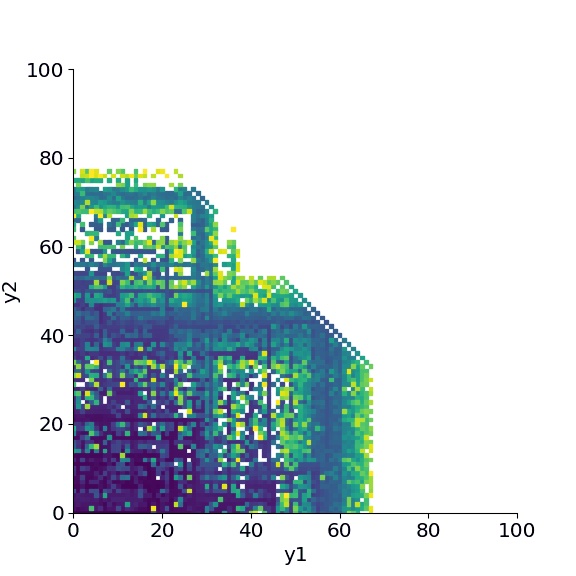}
    \caption{Solutions obtained at different stages of a run of GSEMO for $100$-dimensional \LOTZ. $y1$ indicates the objective of ``LeadingOnes'', and $y2$ indicates the objective of ``Trailing Zeros''.  Darker color indicates that the corresponding solution has been obtained earlier. Complete gif figures are available in~\cite{zenodo}.}
    \label{fig:stages}
\end{figure}

\begin{figure*}
    \centering
    \includegraphics[width=\linewidth]{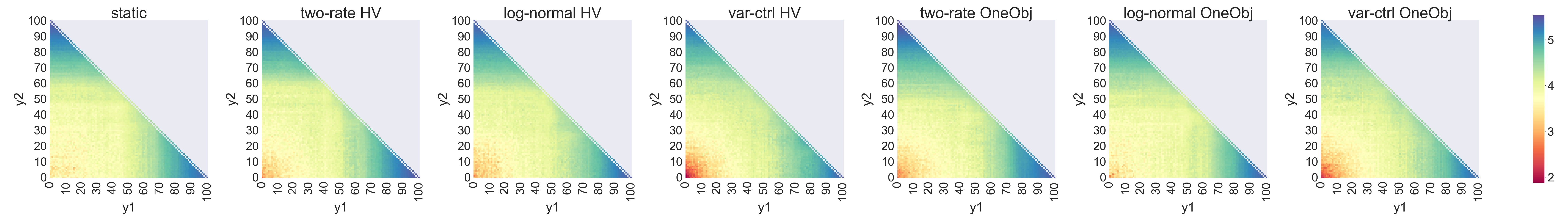}
    \includegraphics[width=\linewidth]{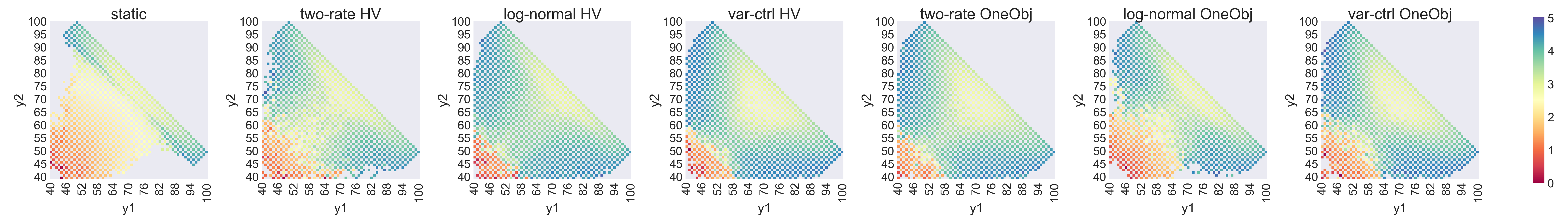}
    \caption{Average function evaluations to find each solution in the search space of $100$-dimensional \LOTZ (top) and \COCZ (bottom). For \LOTZ,
    $y1$ indicates the objective of ``LeadingOnes'', and $y2$ indicates the objective of ``Trailing Zeros''.  For \COCZ, $y1$ indicates the objective of ``OneMax'', and $y2$ indicates the objective of ``Count Ones Count Zeroes'' for each half of the bit-string.  Values are in $\log10$ scale. $\lambda = 10$}
    \label{fig:opt}
\end{figure*}

\subsubsection{Convergence Process}
Figure~\ref{fig:variables} illustrates the convergence process of one objective ($y2$), HV, and the number of obtained Pareto solutions (NUM) for the tested self-adaptive methods. Only the results of the method categories that outperformed the static GSEMO, namely using HV for \OMM and \COCZ and using OneObj for \LOTZ, are displayed. To avoid redundancy, only one of the two single objectives ($y1$ and $y2$) is plotted as they exhibit identical trends.

For \OMM, we observe that the var-ctrl algorithm has an advantage in the beginning, but its performance deteriorates later. Although the two-rate algorithm shows similar behavior to the static GSEMO throughout the optimization process, it requires fewer average function evaluations to obtain the entire Pareto solution set (see Table~\ref{tab:FE}). The optimal mutation rates of MOEAs for \OMM are still unknown, but based on the study of \OM~\cite{GiessenW18,BuzdalovD20}, we intuitively expect that the optimal mutation strength for \OMM is $\ell=1$ at the late stage of the optimization process. Therefore, it is promising that the two-rate algorithm follows the trend of the static GSEMO, as shown in Figure~\ref{fig:variables}-a, that the convergence lines of the two-rate and static GSEMO overlap in the end.

The analysis of the optimization process of SEMO and GSEMO for \LOTZ carried out in~\cite{DBLP:journals/tec/LaumannsTZ04,1299908} can be divided into two phases: \emph{obtaining the first Pareto optimal solution} and \emph{computing the entire Parete front}. 
As shown in Figure~\ref{fig:opt}, the algorithms first hit a solution in the middle of the Pareto front and then use more function evaluations to spread to 
other Pareto solutions. 
The classic runtime analysis~\cite{DBLP:journals/tec/LaumannsTZ04} proves that the running time, i.e., function evaluations, of SEMO is $O(n^2)$ for the first phase as the population contains exactly one individual before having obtained a Pareto optimal solution for the first time.
Furthermore, it proves that SEMO finishes the second phase in exected time $\Theta(n^3)$ by considering the time to compute the remaining Pareto optimal solutions with a population size of at most $n+1$.

The proof for GSEMO given in \cite{1299908} shows that the first Pareto optimal solution is obtained in expected time $O(n^3)$ which works under the assumpting that the population can become of size $\Theta(n)$ before the first Pareto optimal solution has been obtained.
Figure~\ref{fig:stages} visualizes the optimization process of GSEMO for LOTZ. It confirms the increase in population size before obtaining the first Pareto optimal solution. In particular, it shows that multiple non-dominated solutions might exist in the first phase before obtaining the first Pareto solution, and new Pareto solutions can be created from mutating non-Pareto solutions. This practical insight into the behavior of GSEMO indicates potential different paths for the theoretical running time analysis. In terms of the comparison among the tested algorithms, we observe in Figure~\ref{fig:opt} that the self-adaptive methods, especially the var-ctrl algorithm,
obtain advantages in the initial stage of the optimization process.
Figure~\ref{fig:variables}-b also shows that the convergence speed of $y2$ changes significantly after the first Pareto solution is obtained. Based on the results of \LO~\cite{DBLP:conf/cec/YeDB19}, we infer that the var-ctrl algorithm has an advantage in the second phase by controlling the variance of the normal distribution sampling $\ell$.

Figure~\ref{fig:opt} depicts that the first Pareto solution of \COCZ, obtained by the GSEMO variants, also lies in the middle of the Pareto front. However, we do not observe a drastic change in the convergence speed for \COCZ in Figure~\ref{fig:variables}-c. The var-ctrl algorithm exhibits faster convergence in the initial stages, but it performs the worst after the first Pareto solution is obtained. Additionally, Figure~\ref{fig:opt} highlights the intriguing convergence process of the GSEMO variants. Compared to the static GSEMO, the self-adaptive algorithms explore more non-Pareto solutions before attaining the entire Pareto set. Furthermore, the static GSEMO moves along a path of many non-Pareto solutions toward the Pareto front. On the contrary, the self-adaptive methods obtain the first Pareto optimal solution while searching through a lower number of non-Pareto solutions.

\section{A Self-adaptive GSEMO}
The results from using the techniques from single objective optimization (see Section~\ref{sec:exp}) show potential for improvement through the use of self-adaptive mutation rates. The OneObj strategy, which focuses on single objectives, can accelerate the convergence speed for \LOTZ by finding solutions located on the edge of the Pareto front faster. Additionally, self-adaptive methods using HV can obtain the entire Pareto front faster for \OMM and \COCZ, but can also lead to failure due to using HV as the metric, as shown in Figure~\ref{fig:variables}-(a,c), the performance of the var-ctrl GSEMO using HV is good when the used function evaluations are less than $10^3$ but deteriorates later.

To address these issues, we propose a novel GSEMO approach using self-adaptive mutation (AGSEMO). This algorithm applies the normalized standard mutation of var-ctrl for the non-dominated solutions with maximal values for $y1$ or $y2$ and uses log-normal adaptive standard bit mutation for the other non-dominated solutions.

\begin{algorithm2e}
\caption{A GSEMO using Self-Adaptive Mutation}
\label{alg:AMOEA}
\textbf{Input:} initial mutation strength  $r = 1$, $c=0$\;
\textbf{Initialization:} Sample $x \in \{0,1\}^n$ uniformly at random (u.a.r.), evaluate $f(x)$, $s = (x,f(x),r,c)$\;
$P \leftarrow \{s\}$\;
\textbf{Optimization:} \While{not stop condition}{
\For{i = 1,\ldots,$\lambda$}{
Select $s \in P$ u.a.r.\;
\eIf{one objective value of $s$ is the best in $P$}{
Sample $\ell^{(i)} \sim \min\{\text{N}_{>0}(s[r],F^{s[c]}(1-s[r]/n)),n/2\}$}{

$p \leftarrow \big(1+\frac{1-s[r]/n}{s[r]/n}\cdot \exp(0.22\cdot \mathcal{N}(0,1))\big)^{-1}$ \;
Sample $\ell^{(i)} \sim \text{Bin}_{>0}(n,p)$

}
create $y^{(i)} \leftarrow \text{flip}_{\ell^{(i)}}(x)$, and evaluate $f(y^{(i)})$\;
\eIf{$\ell^{(i)} = s[c]$}{$c^{(i)} \leftarrow s[c] +1$}{$c^{(i)} \leftarrow 0$}
}
\For{$i = 1,\ldots,\lambda$}{
$s^{(i)} \leftarrow (y^{(i)},f(y^{(i)}),\ell^{(i)}, c^{(i)})$\;
\If{there is no $z \in P$ such that $s^{(i)}[x] \preceq z[x]$}{$P = \{z \in P \mid z[x] \not\preceq s^{(i)}[x]\} \cup \{s^{(i)}\}$}}
}
\end{algorithm2e}

Algorithm~\ref{alg:AMOEA} presents the proposed AGSEMO in detail. Each solution individual $s$ is represented by its genotype $x$, objective function values $f: \{0,1\}^n \rightarrow \mathbb{R}^2$
, and mutation strength $r$ (line 2). We use $s[*]$ denote the corresponding elements $*$ of $s$. A solution $s$ is randomly selected from the current non-dominated set and mutated (line 6). We consider two different scenarios for mutation. If one objective fitness of $s$ is the maximum among the current non-dominated set (line 7), we apply the normalized bit mutation (line 8). The purpose of this scenario is to optimize for single objectives. In contrast, we use the log-normal strategy to adjust the mutation rate with equal possibilities of increase and decrease (lines 10). If a new non-dominated solution is created, its mutation strength is inherited for future mutation (line 20). 

Note that, for GSEMO, the mutation strength $r$ is embedded with the solution, and its value can be adjusted based on the progress made (e.g., obtaining a new non-dominated solution or not) over time. We apply the var-ctrl strategy to mutate ``edge-solutions'', i.e., those that achieve the best fitness with respect to one of the objectives, because it has shown promising results for optimizing single objectives such as \OM and \LO~\cite{DBLP:conf/cec/YeDB19}. Self-adaptation has been observed to help GSEMO for \OMM, which addresses the phase of spreading out to the entire Pareto front. We apply the log-normal mutation to other non-dominated solutions because it allows us to adjust mutation rates without a pre-defined factor of 
the two-rate strategy. 

The AGSEMO algorithm provides a versatile approach for developing self-adaptive MOEAs. In future applications, the var-ctrl strategy can be replaced by a more specialized technique for specific objectives. Similarly, though the log-normal strategy is already a general approach that has shown to be effective in many problems, it can be substituted with other techniques, such as a local search for each cluster of Pareto solutions.

Table~\ref{tab:AGSEMO} compares the mean and variance of function evaluations that GSEMO and AGSEMO use to obtain the entire Pareto front set of \OMM, \LOTZ, and \COCZ, when using $\lambda = 10$. The results show that AGSEMO outperforms GSEMO across all the problems by using $10\%$, $8\%$, and $11\%$ less function evaluations. The Mann-Whitney U rank test is performed for the running time, i.e., function evaluations, of the two algorithms ($100$ independent runs for each), and the $p$-values are $0.019$, $0.014$, and $0.033$ for \OMM, \LOTZ, and \COCZ, respectively. It is worth mentioning that we present only the results of $\lambda = 10$ in this paper, and our experiment of $\lambda \in \{20,30,40,50\}$, which is available in~\cite{codes,zenodo}, shows consistent comparison results (including those in Section~\ref{sec:exp}). 

\begin{table}[t]
\centering
\begin{tabular}{lrrr}
\toprule
& \OMM & \LOTZ & \COCZ \\
\midrule
AGSEMO & $61\,618$ &$314\,000$ &  $26\,844$ \\
&  (2.938e+08) & (6.365e+09) &  (8.353e+07) \\
\bottomrule
\end{tabular}
    \caption{The mean and variance (in brackets) of function evaluations that AGSEMO uses to obtain the entire Parent front of the $100$-dimensional \OMM, \COCZ, and \LOTZ. The results are average of $100$ runs.}
    \label{tab:AGSEMO}
\end{table}

\section{Conclusions}
Self-adaptive mutation has succeeded in accelerating the convergence speed in single-objective pseudo-Boolean optimization. This paper studies the benefits of self-adaptive mutation in multi-objective optimization, starting with translating single-objective techniques. Detailed experimental investigations provide us insights into the impact of mutation rate for GSEMO. Moreover, we propose a general GSEMO using self-adaptive mutation, presenting competitive results for \OMM, \LOTZ, and \COCZ. 

\emph{Experimental Investigation on GSEMO.}
Our results indicate potential improvement space of GSEMO from using self-adaptive mutation. Hypervolume and the objective-oriented metric are useful to guide GSEMO, but none of the tested metrics is leading across all the tested problems. Meanwhile, the tested adaptive strategies, i.e., two-rate, log-normal, and var-ctrl, also obtained varying performances on the problems. Hypervolume can lead to fast initial convergence of algorithms. However, using hypervolume can also result in slow convergence in the late stages of optimization, but focusing on single objectives can solve this issue.
Such observation inspires us to the proposed self-adaptive AGSEMO. More detailed benchmarking data and visualization of GSEMO variants are available in~\cite{codes,zenodo}. We hope that our practical results can promote novel theoretical insights into MOEAs.

\emph{A Self-adaptive GSEMO.}
The proposed AGSEMO applies different strategies for the solutions with one objective fitness which is the maximum among current non-dominated solutions. In our experiments, we apply the var-ctrl strategy, which performs well for \OM and \LO, for the non-dominated solutions obtaining the best-so-far single objectives. Meanwhile, the log-normal strategy is applied to mutate other non-dominated solutions. The var-ctrl can help move towards the Pareto front faster and explore search space, and the log-normal strategy can adjust mutation rates for mutating non-dominated solutions to fill out the Pareto front. Results show that AGSEMO outperforms the static GSEMO for all the tested problems.

\emph{Future Work.}
Regarding the practical investigation of MOEAs, we have primarily focused on the impact of mutation rates in this work. In our future work, we plan to expand our study for different components and strategies of MOEAs, such as crossover, population size, and selection of non-dominated solutions for variation.
Furthermore, the proposed AGSEMO algorithm is based on the essential idea of using single-objective oriented techniques to explore and proper strategies to exploit search space. This idea can be generalized to other applications such as bi-objective sub-modular optimization. Furthermore, we plan to test this technique on classic many-objective optimization problems, such as mLOTZ and mCOCZ, to examine its scalability and performance.

\bibliographystyle{acm}
\bibliography{references}
\end{document}